\documentclass[10pt,letterpaper]{article}
\usepackage[top=2.25cm,bottom=2.5cm,left=2.5cm,right=2.5cm]{geometry}
\usepackage[T1]{fontenc}
\usepackage[scaled=0.88]{helvet}
\usepackage{microtype}
\usepackage{setspace}
\usepackage{parskip}
\usepackage{hyphenat}
\usepackage{titlesec}
\usepackage[numbers,square,sort&compress]{natbib}
\usepackage{xcolor}
\definecolor{linkblue}{HTML}{0064E0}
\usepackage[breaklinks,colorlinks,allcolors=linkblue]{hyperref}

\usepackage{enumitem}
\usepackage{amsmath}
\usepackage{amsfonts}
\usepackage{bm}
\usepackage{amssymb}
\usepackage{graphicx}
\usepackage{adjustbox}
\usepackage{float}
\usepackage{booktabs}
\usepackage{multirow}
\usepackage{makecell}
\usepackage{caption}
\usepackage{subcaption}
\usepackage{colortbl}
\usepackage[most]{tcolorbox}
\usepackage{xspace}
\usepackage{marvosym}
\usepackage{fontawesome5}

\titleformat*{\section}{\Large\sffamily\bfseries}
\titleformat*{\subsection}{\large\sffamily\bfseries}
\titleformat*{\subsubsection}{\normalsize\sffamily\bfseries}
\DeclareTextFontCommand{\textbf}{\bfseries\sffamily}
\DeclareCaptionLabelSeparator{frontsep}{}
\DeclareCaptionFormat{frontstyle}{{\sffamily\textbf{#1 #2}} #3}
\renewenvironment{abstract}{\par\noindent\ignorespaces}{\par}

\definecolor{exp_table_blue}{HTML}{DAECED}
\definecolor{gray}{gray}{.9}
\definecolor{frontgray}{HTML}{F1F4F7}
\definecolor{fronttext}{HTML}{1C2B33}

\newcommand{\tabincell}[2]{\begin{tabular}{@{}#1@{}}#2\end{tabular}}
\newcommand{\ie}{\emph{i.e.,}\xspace}
\newcommand{\eg}{\emph{e.g.,}\xspace}
\newcommand{\secref}[2][0.5mm]{Sec.\hspace{#1}\ref{#2}\xspace}
\renewcommand{\eqref}[2][0.5mm]{Eq.\hspace{#1}(\ref{#2})\xspace}
\newcommand{\figref}[2][1mm]{Figure\hspace{#1}\ref{#2}\xspace}
\newcommand{\tabref}[2][1mm]{Table\hspace{#1}\ref{#2}\xspace}

\newcommand{\gainred}[1]{\textcolor[HTML]{D32600}{\textbf{{#1}}}}

\begin{document}
\begin{tcolorbox}[
  enhanced,
  colback=frontgray,
  frame hidden,
  arc=10pt,
  left=0.5cm,right=0.5cm,top=0.5cm,bottom=0.5cm,
  before skip=0pt,
  grow to left by=1.5pt,grow to right by=1.5pt
]
\setlength{\parindent}{0pt}
\setlength{\parskip}{0.5cm}
\begingroup
\setlength{\parskip}{0pt}
\raggedright\nohyphens
{\setstretch{1.618}
{\huge\sffamily\bfseries Unlocking Token Rewards via Training-Free Reward Attribution}\par}
\vskip 0.2cm
{\sffamily\bfseries
Sitong Wu$^{1}$, Haoru Tan$^{3,\dagger}$, Bin Xia$^{1}$, Xichen Zhang$^{2}$,
Jingyao Li$^{1}$, Shaofeng Zhang, Xiaojuan Qi$^{3}$, Bei Yu$^{1}$,
Jiaya Jia$^{1,2,\dagger}$\par}
\vskip 0.2cm
{\normalsize
$^{1}$Chinese University of Hong Kong, \quad $^{2}$Hong Kong University of Science and Technology\par
$^{3}$University of Hong Kong\par}
\endgroup
{\color{fronttext}\begin{abstract}%
In this paper, we propose an extremely efficient, training-free method to extract token-level reward signals directly from an existing deep reward model. Our core idea is to attribute the overall process reward to individual tokens by estimating each token's influence. This influence is defined as the change in the final macroscopic reward (e.g., the process reward) when a token is replaced with a semantically null token. Naively calculating this influence is computationally infeasible, requiring $N$ forward passes through the PRM for an $N$-token sequence. We overcome this bottleneck by proposing a highly efficient gradient-based estimator. Specifically, we use a first-order Taylor approximation, which simplifies the influence calculation to the inner product of the difference between the token embedding and the null token embedding, and the gradient of the reward with respect to the token embedding. This requires only a single forward and backward pass. The resulting token-level rewards enable standard RL algorithms to perform precise credit assignment without requiring additional reward model training. Experiments on challenging reasoning benchmarks demonstrate that our method substantially improves policy optimization efficiency and enhances the generalization of LLM reasoning capabilities. Our P2T outperforms the outcome reward by +4.9\% on MathVista for Qwen2.5-VL-7B-Instruct, and +11.5\% on AIME24 for Qwen2.5-Math-7B.
\end{abstract}

\par}
\vskip 0.5cm
\begingroup
\setlength{\parskip}{0pt}
\small
{\sffamily\bfseries Conference:} CVPR 2026\par
{\sffamily\bfseries Last updated:} \today\par
{\sffamily\bfseries Code:} \url{https://github.com/stonewst/P2T}\par
\endgroup
\end{tcolorbox}

\begingroup
\renewcommand{\thefootnote}{\fnsymbol{footnote}}
\footnotetext[2]{Corresponding author. \par  \faEnvelope\ Sitong Wu: \texttt{stone-wu@link.cuhk.edu.hk} and \texttt{stonewst@163.com}}
\endgroup

\section{Introduction}

Reinforcement Fine-Tuning (RFT) has become a cornerstone paradigm for advancing the reasoning capabilities of Large Models \citep{deepseek-math,deepseek-r1,gemini-flash,o1,open-reasoner-zero}. The two most prevalent forms of supervision signals are the Outcome Reward (based on the final answer's correctness) and the Process Reward. The latter typically requires training a dedicated Process Reward Model (PRM) to evaluate the quality of individual steps within a reasoning trajectory. 
However, both outcome and process rewards ultimately share the limitation of coarse-grained supervision. Even PRM step-level scores remain ambiguous across the hundreds or thousands of tokens in a single step.

In parallel, recent work attempts to decompose coarse rewards into fine-grained, token-level supervision. Existing methods fall into two categories: 
The first involves training a dedicated token-level reward model, often relying on weakly- or unsupervised methods \citep{prime,oreal} to generate synthetic labels. However, this approach incurs substantial training costs while yielding reward models of uncertain reliability due to the high noise and low fidelity of the pseudo-labels. The second avenue employs cheap, heuristic proxy metrics, such as token entropy \citep{emft}, as substitutes for true rewards. While computationally inexpensive, these proxies fundamentally lack semantic alignment with the actual token quality; they serve as mere heuristics only loosely correlated with a token's contribution to the final reward, rather than principled reward estimators. Consequently, current token-level supervision is caught in a dilemma: either incur heavy costs for uncertain models or compromise optimization fidelity with inaccurate proxy signals. Crucially, training a truly fine-grained, token-level reward model is infeasible due to the prohibitive expense of acquiring large-scale human annotations. 
A critical open question emerges: \textit{How can we devise a method for token-level reward estimation that is simultaneously cost-effective, principled, and robust?}

In this paper, we propose a principled, efficient, and robust solution, termed Process-to-Token $\textbf{(P2T)}$ reward attribution. P2T directly decomposes the reward from any differentiable coarse-grained reward model (like a PRM) into accurate token-level rewards. 
Our core idea is to attribute the influence of each token on the coarse-grained reward. Specifically, a token's influence is defined as the change in the overall process reward if that token were replaced with a non-informative ``null'' token. However, a naive implementation of this approach would require $N$ separate PRM inferences (one per token) for an $N$-token sequence, rendering it computationally expensive. To achieve efficient computation, we propose an approximation solver. This estimator allows us to obtain the token-level influence score for all tokens in the entire process simultaneously with only a single forward and backward pass. Conclusively, the contribution of any token to the whole process is approximated using a first-order Taylor expansion: the inner product between (a) the gradient of the process reward with respect to the token's embedding, and (b) the difference between the token's original embedding and the null token embedding. 
Subsequently, we normalize the calculated token influences and use them as weights to redistribute the coarse-grained process reward, yielding the final token-level reward for each token.

Our method offers three key advantages:
(1) \textbf{Training-free}.
It builds on the powerful PRM while eliminating the significant cost of training dedicated token reward models.
(2) \textbf{High fidelity}.
The resulting token rewards are attributed based on reliable PRM outputs, significantly outperforming proxy metrics in both interpretability and accuracy.
(3) \textbf{Scalable efficiency}.
The gradient-based approximation enables the efficiency of our attribution computation.
Extensive experiments demonstrate that our P2T significantly improves policy optimization stability and final reasoning accuracy across various benchmarks, outperforming both proxy-based and trainable token reward baselines. 
For example, P2T outperforms the outcome reward by +4.9\% on MathVista for Qwen2.5-VL-7B-Instruct. For text-only LLMs, P2T still exhibits strong advantages of +11.5\% on AIME24 over the outcome reward for Qwen2.5-Math-7B and +14.2\% on MinervaMath for DeepSeek-R1-Distill-Qwen-1.5B.
In addition, P2T improves training efficiency, reaching the best test accuracy of the outcome-reward counterpart with 3--4$\times$ fewer training steps.
Our contributions are threefold:
\begin{itemize}[leftmargin=0.4cm,topsep=0pt,itemsep=0pt]
\item We propose a novel and principled algorithm to convert process-level rewards into token-level rewards in a completely training-free manner, effectively bridging the gap between coarse and fine-grained supervision for reinforcement learning.
\item We propose a highly efficient approximation for token contribution attribution, reducing the complexity from $\mathcal{O}(N)$ to $\mathcal{O}(1)$, ensuring its practical application to large-scale RL training.
\item We empirically demonstrate that using the derived token-level rewards for RL training leads to significant improvements in the reasoning performance of LLMs compared to both sparse rewarding methods and other fine-grained rewarding methods.
\end{itemize}

\section{Method}

In this section, we describe our approach to converting the more macroscopic rewards provided by a reward model into token-level rewards in an efficient, training-free manner. The key idea is token attribution, that is,  the marginal effect of each token on the overall reward. 
Then we will show how to estimate this using first-order gradient information, thereby enabling reinforcement fine-tuning with token-level credit assignment.

\vspace{0.1cm}
\begin{figure}[!t]
\centering
\includegraphics[width=1.0\linewidth]{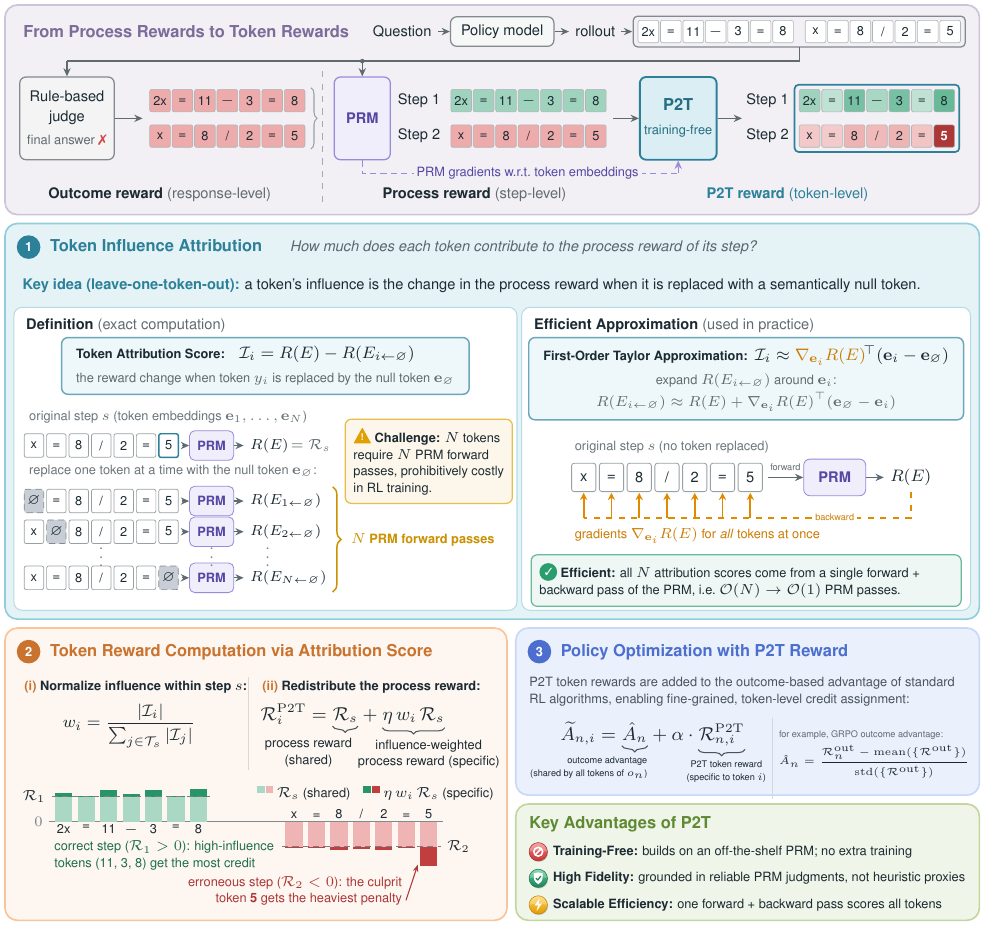}
\vspace{-0.4cm}
\caption{
Overview of Process-to-Token (P2T) reward attribution. P2T converts the process-level rewards provided by a Process Reward Model (PRM) into fine-grained token-level rewards. Specifically, P2T first evaluates the specific influence (attribution score) of each token on the overall process reward, and then normalizes these scores to serve as weights for redistributing the original process reward among the tokens. Consequently, highly influential tokens in a correct step receive the most credit, while the ``culprit'' tokens responsible for an erroneous step incur the heaviest penalty. Finally, these fine-grained token-level rewards are directly incorporated into standard RL algorithms to enable precise credit assignment.
}
\vspace{0.4cm}
\end{figure}

\subsection{Definition of Token Attribution Score}

The output of the policy model is treated as a sequence of tokens $Y = (y_1, y_2, \ldots, y_N)$, where each token $y_i$ is associated with an embedding vector $\mathbf{e}_i \in \mathbb{R}^d$ in the reward model. These vectors form the full sequence embedding $E = [\mathbf{e}_1, \mathbf{e}_2, \ldots, \mathbf{e}_N]$. A differentiable reward model $R(\cdot)$ then computes a scalar reward $R(E) \in \mathbb{R}$ for the entire sequence. 
We define the \textbf{Token Attribution Score} $\mathcal{I}_i$ for a token $y_i$ as the precise difference in the total reward when its embedding $\mathbf{e}_i$ is replaced by a dedicated \emph{null token} embedding $\mathbf{e}_{\varnothing}$, without changing the sequence length:
\begin{equation}
\begin{aligned}
    \mathcal{I}_i = R(E) - R(E_{i \leftarrow \varnothing}),
\end{aligned}
\end{equation} 
where $E_{i \leftarrow \varnothing}$ denotes the modified sequence embedding where the $i$-th token's embedding $\mathbf{e}_i$ is substituted by $\mathbf{e}_{\varnothing}$, $R(E)$ is the original reward, and $R(E_{i \leftarrow \varnothing})$ is the new reward after null-token replacement. 
This score provides a direct, localized measure of the marginal importance of token $y_i$ to the final sequence reward. A positive score ($\mathcal{I}_i > 0$) indicates that $y_i$ was a positive contributor to the sequence's high reward, as replacing it caused the reward to drop. Conversely, a negative score ($\mathcal{I}_i < 0$) suggests $y_i$ was a detrimental contributor that lowered the overall reward, and a score near zero signifies that $y_i$ has a negligible marginal effect.

\noindent\textbf{Choice of NULL Tokens.} We select the padding token from the reward model's existing vocabulary to serve as the null token $\mathbf{e}_{\varnothing}$ for calculating the Token Attribution Score $\mathcal{I}_i$. Although the padding token is not necessarily an untrained token (\eg \texttt{<|endoftext|>} in Qwen-based models also marks document boundaries during pre-training), it virtually never appears within a reasoning response and therefore carries little content-level meaning in this context. Substituting the $i$-th token's embedding $\mathbf{e}_i$ with $\mathbf{e}_{\varnothing}$ thus approximates removing the content of token $y_i$, while keeping the input a real token embedding that the reward model is familiar with, rather than an artificial vector. The resulting attribution score $\mathcal{I}_i = R(E) - R(E_{i \leftarrow \varnothing})$ then quantifies the reward change incurred when the contribution of token $y_i$ is removed. We compare alternative choices of the null token in \secref{sec:ablation}.

\subsection{Efficient Approximation}

While conceptually simple and direct, this substitution method is computationally intensive, requiring $N$ forward passes through the reward model to compute all token scores for a sequence of length $N$, which motivates the search for efficient gradient-based approximations.

We approximate it via a first-order Taylor expansion around $\mathbf{e}_i$:
\begin{equation}
\begin{aligned}
\mathcal{I}_i \approx\; \nabla_{\mathbf{e}_i} R(E)^\top \left(\mathbf{e}_i - \mathbf{e}_\varnothing \right).
\label{eq:taylor}
\end{aligned}
\end{equation}
Here, $\nabla_{\mathbf{e}_i} R(E)$ is the gradient of the reward with respect to the embedding of token $y_i$. This approximation can be computed efficiently in a single backward pass, since gradients with respect to all token embeddings are obtained simultaneously. 
A positive value indicates that token $y_i$ contributes positively to the reward, while a negative value suggests the opposite.

\subsection{Process-to-Token Reward Attibution} 

Next, we will introduce some application details: how to calculate our P2T token reward in \secref{sec:p2t_cal}, and how to integrate it with popular RL algorithms in \secref{sec:p2t_with_grpo}.

\subsubsection{P2T Token Reward Calculation}
\label{sec:p2t_cal}
The attribution score $\mathcal{I}_i$, derived from the gradient-based approximation in \eqref{eq:taylor}, provides a fine-grained token influence score. 
In practice, a PRM evaluates a reasoning trajectory step by step. Let $\mathcal{T}_s$ denote the set of tokens in the $s$-th reasoning step, and $E_{\le s}$ the embeddings of all tokens up to the end of this step. For brevity, we omit the index of the response in this subsection. Since the PRM is causal, the process reward of the $s$-th step is given by $\mathcal{R}_s = R(E_{\le s})$, and the attribution score $\mathcal{I}_i$ of each token $y_i \in \mathcal{T}_s$ is computed by \eqref{eq:taylor} with $R(E)$ instantiated as $R(E_{\le s})$.

Rather than using these scores directly as rewards, we normalize the calculated token influences within each step and use them as weights to redistribute the coarse-grained process reward $\mathcal{R}_s$, yielding the final token-level reward $\mathcal{R}^{\text{P2T}}_i$ for each token $y_i \in \mathcal{T}_s$: 
\begin{equation}
\mathcal{R}^{\text{P2T}}_i = \mathcal{R}_s + \eta \cdot w_i \cdot \mathcal{R}_s, \qquad w_i = \frac{|\mathcal{I}_i|}{\sum_{j \in \mathcal{T}_s} |\mathcal{I}_j|},
\label{eq:p2t_reward}
\end{equation} 
where $\eta$ controls the strength of the token-level redistribution.
The magnitude $|\mathcal{I}_i|$ measures how strongly token $y_i$ affects the process reward, and the normalized weight $w_i$ is its share of the total influence within the step. Hence, the most influential tokens of a correct step ($\mathcal{R}_s > 0$) receive the largest rewards, while the most influential tokens of an erroneous step ($\mathcal{R}_s < 0$) receive the heaviest penalties.
This ensures that the reward $\mathcal{R}_s$ is distributed according to the measured marginal contribution of each token, allowing the policy to accurately learn which specific actions (token generation) led to the high or low coarse-grained reward. 

It is worth noting that we do not entirely discard the coarse-grained process reward $\mathcal{R}_s$ in this framework. In the calculation of our final token-level reward $\mathcal{R}^{\text{P2T}}_i$, we introduce $\mathcal{R}_s$ as a baseline (the first term in the equation), and the estimated token-level influence scores, after normalization, are reweighted by a factor $\eta$ and added to this baseline to yield the final token reward. The coarse-grained process reward $\mathcal{R}_s$ serves as a baseline, providing a stable reward signal to every token in the step. This is crucial because the attribution scores $\mathcal{I}_i$, often derived from approximate gradient calculations, can be inherently noisy or imprecise. Introducing $\mathcal{R}_s$ as the baseline limits the disturbance that attribution errors can introduce to the final reward, preventing the policy from fluctuating too wildly or being dominated by a few noisy token rewards during the learning phase. 

\noindent\textbf{Magnitude vs.\ signed influence.} Since the sign of $\mathcal{I}_i$ indicates whether token $y_i$ raises or lowers the process reward, a natural alternative is a signed redistribution that rewards the tokens pushing the step reward up and penalizes those pulling it down:
\begin{equation}
\mathcal{R}^{\text{P2T-signed}}_i = \mathcal{R}_s + \eta \cdot \operatorname{sign}(\mathcal{I}_i) \cdot w_i \cdot |\mathcal{R}_s|.
\label{eq:p2t_reward_signed}
\end{equation}
where $\operatorname{sign}(\cdot)$ denotes the sign function, which returns $1$, $-1$, and $0$ for positive, negative, and zero inputs, respectively.
However, we empirically find that this signed variant underperforms the magnitude-based reward in \eqref{eq:p2t_reward} (see Table~\ref{tb:exp_ablation}(c)). One possible reason is that the positive and negative terms in \eqref{eq:p2t_reward_signed} are generally imbalanced within a step. We therefore adopt \eqref{eq:p2t_reward} by default, and leave a more effective use of the sign information to future work.

\subsubsection{GRPO with P2T Token Reward}
\label{sec:p2t_with_grpo}
The core step in applying token-level rewards to standard reinforcement learning algorithms is correctly calculating the advantage function at the token level. Taking GRPO \citep{deepseek-math} as an example, it is designed to be value-function-free and estimates the advantage by comparing the reward of a generated response to the average reward of other responses in the same group. 
\begin{equation}
\hat{A}_{n} = \frac{\mathcal{R}^{\text{out}}_{n} - \text{mean}(\{\mathcal{R}^{\text{out}}\})}{\text{std}(\{\mathcal{R}^{\text{out}}\})},
\end{equation}
where $\mathcal{R}^{\text{out}}_{n}$ is the outcome reward of $n$-th response. For example, for math problems, it can be the correctness indicator of the final answer ($1$ for correct and $-1$ for wrong).
When incorporating our P2T token-level rewards, the advantage will be 
\begin{equation}
\widetilde{A}_{n,i} = \hat{A}_{n} + \alpha \cdot \mathcal{R}^{\text{P2T}}_{n,i},
\label{eq:p2t_adv}
\end{equation}
where $i$ is the token index, and $\mathcal{R}^{\text{P2T}}_{n,i}$ denotes the P2T reward for $i$-th token in $n$-th response. $\alpha$ is a weight to balance the contribution of the original outcome-derived advantage and our token-level guidance. By default, we set $\alpha=1.0$. When the PRM is less reliable, a smaller $\alpha$, or a schedule that gradually increases $\alpha$ during training, can help stabilize optimization.

\section{Experiments}
\label{sec:exp}

\subsection{Settings}

\noindent\textbf{Baseline Models.}
We conduct RL training on various multimodal and text-only models, encompassing a diverse range of models from the Qwen and LLaMA series with varying sizes (from 1.5B to 7B).
For multimodal models, we utilize Qwen2.5-VL-7B-Instruct \citep{qwen2.5-vl} for its popularity and strong performance. 
For text-only models, we consider both base models after pre-training (Qwen2.5-Math \citep{qwen2.5-math}), and instruction-tuned models from post-training with different reasoning paradigms: CoT model (LLaMA3.2-Instruct \citep{llama3}), LongCoT models (DeepSeek-R1-Distill-Qwen \citep{deepseek-r1}, DeepScaleR-1.5B-Preview \citep{deepscaler}), and hybrid-mode models (Qwen3 \citep{qwen3}). 
We additionally use Eurus-2-7B-SFT \citep{prime} for comparison with existing token-level rewards, Vision-R1-7B \citep{vision-r1} and DeepScaleR-1.5B-Preview \citep{deepscaler} for continued training, and Qwen2.5-Math-1.5B \citep{qwen2.5-math} for analyzing training dynamics, where Qwen2.5-Math-1.5B is trained under exactly the same setting as Qwen2.5-Math-7B.

\noindent\textbf{Training Dataset.}
For multimodal LLMs, we employ a mixture of two data sources: VerMulti-65K \citep{lmm-r1} and Vision-R1-10K \citep{vision-r1}, ensuring a coverage of both general and math multimodal tasks.
For text-only LLMs, the training data is specialized by different models and tasks. By default, we use the MATH \citep{math} dataset for mathematical training.
For recent models, like DeepScaleR-1.5B-Preview \citep{deepscaler} and Qwen3 \citep{qwen3}, we utilize the more challenging DeepScaleR-40K \citep{deepscaler} as training data accordingly.
For the Eurus-2-7B-SFT \citep{prime} model, we adopt the Eurus-2-RL-Data \citep{prime} as its training corpus, following prior works \citep{prime,spro} for a fair comparison.

\noindent\textbf{Evaluation.}
All the multimodal LLMs are evaluated on three multimodal math benchmarks (MathVista \citep{mathvista}, MathVerse \citep{mathverse}, and MathVision \citep{mathvision}) and two general multimodal benchmarks (MMMU \citep{mmmu} and MMStar \citep{mmstar}).
For text-only LLMs trained on math data, the evaluation is conducted on seven commonly-used challenging math benchmarks, including AIME24 \citep{aime}, AIME25 \citep{aime}, AMC23 \citep{amc23}, MATH-500 \citep{math}, MinervaMath \citep{minerva}, OlympiadBench \citep{olympiadbench}, and GaoKao2023en \citep{gaokao2023en}.
Following \citep{prime,spro}, for the experiments on Eurus-2-7B-SFT \citep{prime} model trained on a combination of math and code data, we perform evaluation on six benchmarks, including AMC23, MATH, OlympiadBench, K12, CodeForces \citep{codeforces}, and CodeContests \citep{codecontests}.
For each test problem, we sample 8 responses with a temperature of 0.6 and a top-p of 0.95, and report the average Pass@1.

\noindent\textbf{Training Settings.}
The multimodal models are trained for 1 epoch with a global batch size of 128. 
The text-only models are trained for 500 $\sim$ 1000 steps with a global batch size of 256. 
We use a constant learning rate of $1 \times 10^{-6}$ and weight decay of 0.01. For each training data, we generate 8 responses with temperature of 1.0 and topp of 1.0. The KL coefficient is set to 0.001.

\noindent\textbf{Implementation Details in P2T.}
Our Process-to-Token (P2T) method relies on a Process Reward Model (PRM) to provide the process-level reward for each reasoning step. 
The choice of the PRM is critical and tailored to the domain. In our experiments, we use VisualPRM \citep{visualprm} to provide the process reward for multimodal models, due to the strength of VisualPRM in handling a diverse range of multimodal tasks, including both general multimodal understanding and domain-specific tasks such as multimodal mathematics reasoning.
For text-only models, we obtain the process reward from ReasonFlux-PRM \citep{reasonflux-prm} as it offers the crucial advantage of supporting and evaluating LongCoT reasoning trajectories, which is not well-supported in earlier PRMs.
For the hyperparameter $\eta$ in \eqref{eq:p2t_reward}, we set $\eta=0.5$ by default. For $\alpha$ in \eqref{eq:p2t_adv}, we set $\alpha=1.0$ for all models.

\vspace{0.6cm}
\begin{table*}[!h]
\centering
\caption{\label{tb:exp_compare_outcome_llm}
Comparison of LLMs trained with GRPO and GRPO+P2T on math reasoning data. Gains of GRPO+P2T over GRPO are highlighted in red. For Qwen3 model, the results are reported in the format: performance in ``thinking mode'' / ``non-thinking mode''.
}
\vspace{-0.1cm}
\setlength\tabcolsep{4pt}
\resizebox{1.0\textwidth}{!}{
\begin{tabular}{l|ccccccc|c}
\toprule[1pt]
\textbf{Model} & \textbf{AIME24} & \textbf{AIME25}  & \textbf{AMC23} & \textbf{MATH-500} & \textbf{MinervaMath}  & \textbf{OlympiadBench} & \textbf{GaoKao2023en} & \textbf{Avg} \\
\midrule
Qwen2.5-Math-7B 
& 13.2 & 9.8 & 40.0 & 53.6 & 17.2 & 17.4 & 39.7 & 27.3 \\
\quad + GRPO
& 28.8 & 14.1 & 67.3 & 81.8 & 36.1 & 41.9 & 65.9 & 48.0 \\
\rowcolor{exp_table_blue}
\quad + GRPO + P2T
& \textbf{40.3} (\gainred{+11.5})  & \textbf{20.0} (\gainred{+5.9})  & \textbf{72.8} (\gainred{+5.5}) & \textbf{84.5} (\gainred{+2.7}) & \textbf{41.9} (\gainred{+5.8}) & \textbf{45.6} (\gainred{+3.7}) & \textbf{71.8} (\gainred{+5.9}) & \textbf{53.8} (\gainred{+5.8}) \\
\midrule
LLaMA3.2-3B-Instruct   
& 3.3  & 0.2 & 22.5 & 48.0 & 16.5 & 14.5 & 33.7 & 19.8 \\
\quad + GRPO        
& 13.3 & 1.5 & 27.5 & 56.8 & 20.2 & 21.4 & 43.8 & 26.4 \\
\rowcolor{exp_table_blue}
\quad + GRPO + P2T  
& \textbf{21.7} (\gainred{+8.4})  & \textbf{10.8} (\gainred{+9.3})  & \textbf{46.5} (\gainred{+19.0}) & \textbf{62.6} (\gainred{+5.8}) & \textbf{27.9} (\gainred{+7.7}) & \textbf{26.3} (\gainred{+4.9}) & \textbf{54.5} (\gainred{+10.7}) & \textbf{35.8} (\gainred{+9.4}) \\
\midrule
DeepSeek-R1-Distill-Qwen-1.5B   
& 28.9  & 22.8 & 62.9 & 83.9 & 26.5 & 43.3 & 73.1 & 48.8 \\
\quad + GRPO
& 30.0  & 23.6 & 67.5 & 84.3 & 27.2 & 45.4 & 74.8 & 50.4 \\
\rowcolor{exp_table_blue}
\quad + GRPO + P2T  
& \textbf{37.8} (\gainred{+7.8})  & \textbf{28.6} (\gainred{+5.0})  & \textbf{76.5} (\gainred{+9.0}) & \textbf{88.2} (\gainred{+3.9}) & \textbf{41.4} (\gainred{+14.2}) & \textbf{49.7} (\gainred{+4.3}) & \textbf{78.4} (\gainred{+3.6}) & \textbf{57.2} (\gainred{+6.8}) \\
\midrule
Qwen3-1.7B 
& 46.7 / 13.4 & 35.4 / 8.8 & 85.3 / 50.5 & 91.9 / 73.0 & 49.2 / 32.6 & 56.4 / 39.2 & 78.3 / 62.4 & 63.3 / 40.0 \\
\quad + GRPO
& 53.7 / 23.5 & 42.5 / 19.5 & 88.7 / 63.1 & 92.8 / 83.5 & 50.7 / 40.6 & 58.0 / 40.3 & 80.5 / 63.9 & 66.7 / 47.8 \\
\rowcolor{exp_table_blue}
\quad + GRPO + P2T
& \tabincell{c}{\textbf{59.0} / \textbf{31.9}\\(\gainred{+5.3}) / (\gainred{+8.4})} & \tabincell{c}{\textbf{47.0} / \textbf{30.1}\\(\gainred{+4.5}) / (\gainred{+10.6})} & \tabincell{c}{\textbf{90.9} / \textbf{70.4}\\(\gainred{+2.2}) / (\gainred{+7.3})} & \tabincell{c}{\textbf{93.4} / \textbf{86.9}\\(\gainred{+0.6}) / (\gainred{+3.4})} & \tabincell{c}{\textbf{52.0} / \textbf{44.2}\\(\gainred{+1.3}) / (\gainred{+3.6})}  & \tabincell{c}{\textbf{59.3} / \textbf{41.8}\\(\gainred{+1.3}) / (\gainred{+1.5})}  & \tabincell{c}{\textbf{82.9} / \textbf{65.6}\\(\gainred{+2.4}) / (\gainred{+1.7})}  & \tabincell{c}{\textbf{69.2} / \textbf{53.0}\\(\gainred{+2.5}) / (\gainred{+5.2})}   \\
\bottomrule[1pt]
\end{tabular}%
} 
\end{table*}
    
\begin{figure}[!h] 
\centering  
\includegraphics[width=1.0\linewidth]{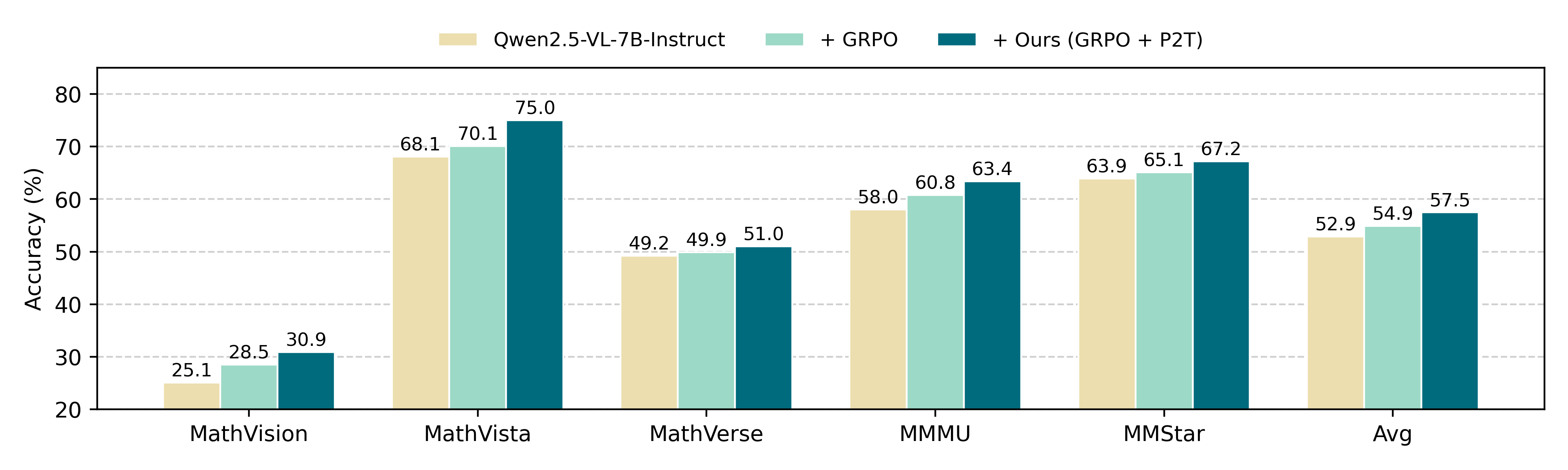}
\vspace{-0.6cm}
\caption{\label{fig:exp_compare_outcome_mllm}
Comparison of Qwen2.5-VL-7B-Instruct trained with GRPO and GRPO+P2T.}  
\end{figure}

\subsection{Main Results}

\subsubsection{Comparison with Outcome Reward RL}

We first compare P2T with the standard outcome reward on both multimodal and text-only LLMs. For multimodal LLMs, we train Qwen2.5-VL-7B-Instruct and report the results in \figref{fig:exp_compare_outcome_mllm}. For text-only LLMs, we cover four reasoning paradigms: a base model (Qwen2.5-Math-7B), a CoT instruct model (LLaMA3.2-3B-Instruct), a LongCoT model (DeepSeek-R1-Distill-Qwen-1.5B), and a hybrid-mode model (Qwen3-1.7B, evaluated in both thinking and non-thinking modes), with results in \tabref{tb:exp_compare_outcome_llm}. For a fair comparison, for each model, GRPO \citep{deepseek-math} and GRPO+P2T are trained under identical settings, and the only difference lies in the reward: GRPO uses the outcome reward alone, while GRPO+P2T additionally incorporates our P2T token reward.

\vspace{0.1cm}
\noindent\textbf{Consistent improvements across models, paradigms, and modalities.}
GRPO+P2T outperforms GRPO on every benchmark for every model: across the 40 model--benchmark pairs (5 for the multimodal model and 35 for the text-only settings), there is not a single exception. The average gains range from +2.5\% to +9.4\%, and the improvements hold for both the Qwen and LLaMA families, for model sizes from 1.5B to 7B, for base, CoT, LongCoT, and hybrid-mode models, and for both text-only and multimodal inputs. This suggests that the token-level credit derived from a PRM is broadly applicable, rather than tied to a particular model family, reasoning format, or modality.

\vspace{0.1cm}
\noindent\textbf{Larger gains where outcome-based RL stalls.}
P2T is most helpful precisely where the outcome reward provides little learning signal. On DeepSeek-R1-Distill-Qwen-1.5B, GRPO improves the average accuracy by only +1.6\% over the initial model, whereas GRPO+P2T improves it by +8.4\%. Similarly, on MathVista, GRPO brings only +2.0\% to Qwen2.5-VL-7B-Instruct, while GRPO+P2T brings +6.9\%. An outcome reward assigns the same credit to every token of a trajectory, so correct and erroneous steps are reinforced or penalized together; once the easy gains are exhausted, this coarse signal makes little further progress. P2T breaks this tie by distinguishing the tokens that actually drive the reward from the rest.

\vspace{0.1cm}
\noindent\textbf{Larger gains on harder benchmarks.}
The improvements from P2T are concentrated on competition-level problems. Averaged over AIME24, AIME25, and AMC23, GRPO+P2T improves over GRPO by +4.0\% to +12.2\% across the five text-only settings, roughly two to seven times its gain on MATH-500 (+0.6\% to +5.8\%). Harder problems require longer chains of multi-step reasoning, in which a single faulty step can invalidate the final answer, making precise credit assignment all the more valuable. Part of this gap may reflect that MATH-500 is close to saturation for most models. However, the gap persists for LLaMA3.2-3B-Instruct, whose MATH-500 accuracy (56.8\% after GRPO) is far from saturated, suggesting that problem difficulty, rather than the remaining headroom alone, drives the larger gains.

\begin{figure}[t] 
\centering
\includegraphics[width=1.0\linewidth]{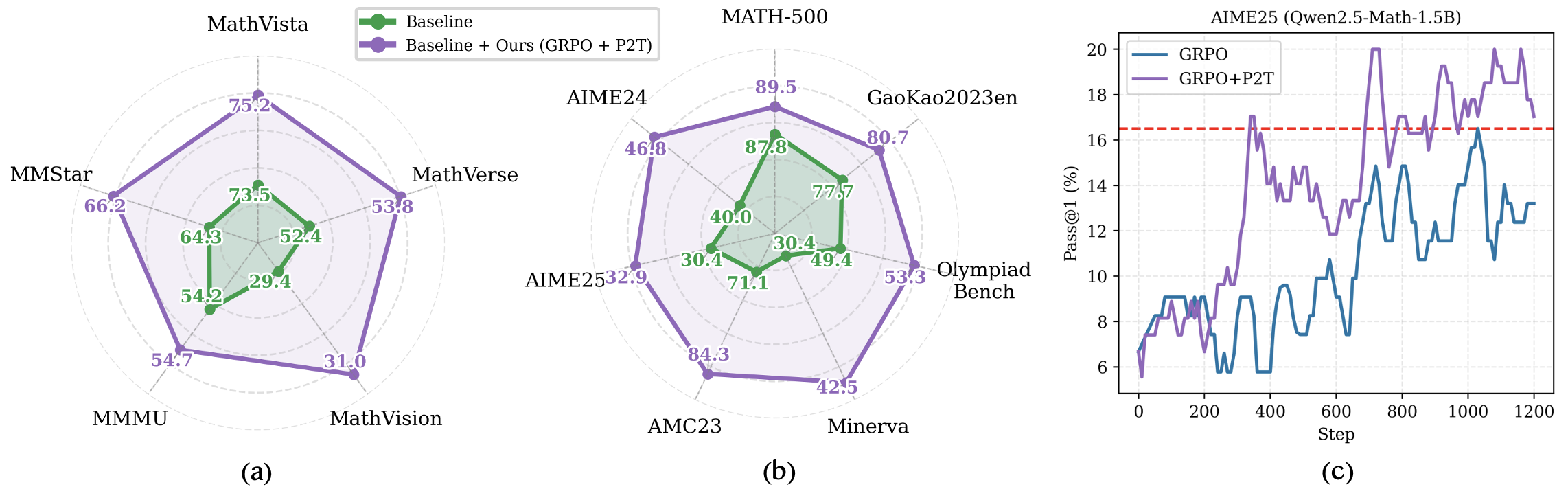}
\vspace{-0.8cm}
\caption{\label{fig:exp_bottleneck_and_dynamic}
P2T raises the performance ceiling and improves the training efficiency of outcome-based GRPO. (a, b) Results of further training two models that have already been trained to convergence with outcome-based RL, (a) Vision-R1-7B \citep{vision-r1} and (b) DeepScaleR-1.5B-Preview \citep{deepscaler}, using GRPO+P2T on their original training data (\ie Vision-R1-10K and DeepScaleR-40K, respectively). (c) AIME25 Pass@1 of Qwen2.5-Math-1.5B during training with GRPO and GRPO+P2T. The dashed line marks the best accuracy reached by GRPO.
}
\vspace{0.4cm}
\end{figure}

\vspace{0.1cm}
\noindent\textbf{Breaking the convergence bottleneck of outcome-based RL.}
P2T can further serve as a second-stage optimization that unlocks performance beyond what outcome-based RL converges to. We take Vision-R1-7B \citep{vision-r1} and DeepScaleR-1.5B-Preview \citep{deepscaler}, two models that have already been fully trained with outcome-based RL, and continue to train them with GRPO+P2T on their original training data (\ie Vision-R1-10K and DeepScaleR-40K, respectively). As shown in \figref{fig:exp_bottleneck_and_dynamic}(a) and (b), both models improve on every benchmark, \eg +6.8\% on AIME24 and +12.1\% on MinervaMath for DeepScaleR-1.5B-Preview. Since no new data is introduced, these gains can be largely attributed to the finer-grained credit assignment provided by P2T.

\vspace{0.1cm}
\noindent\textbf{Higher training efficiency.}
Beyond the final accuracy, P2T also makes training more efficient. \figref{fig:exp_bottleneck_and_dynamic}(c) tracks the AIME25 accuracy of Qwen2.5-Math-1.5B throughout training. GRPO+P2T surpasses the best accuracy of GRPO (dashed line) much earlier than GRPO itself reaches this level, \ie P2T reaches the same test performance with 3--4$\times$ fewer training steps. Moreover, GRPO+P2T stays well above GRPO for the rest of training and ends with a clear margin.

\subsubsection{Comparison with Existing Token-level Reward}

We further compare P2T with two representative token-level reward methods: PRIME \citep{prime}, which trains an implicit process reward model online alongside the policy and derives token-level rewards from it, and SPRO \citep{spro}, which derives token-level rewards from the policy model itself via its log-probability ratio to the reference model. Following \citep{prime}, all methods start from the same Eurus-2-7B-SFT model and are trained on Eurus-2-RL-Data. As shown in \tabref{tb:exp_compare_dense_reward}, P2T achieves the best result on all six benchmarks, improving the average accuracy over GRPO by +7.2\%, over PRIME by +4.7\%, and over SPRO by +2.3\%.

These gains stem from how the token-level credit is obtained and what it measures. PRIME has to update its reward model during RL using only outcome labels, so the quality of its token-level credit depends on how well this model is learned on the fly. P2T instead extracts token-level credit from an off-the-shelf PRM without any additional training, and introduces no extra model to optimize during RL. SPRO avoids such a learned reward model, but its reward reflects how the policy deviates from the reference model, which is not a direct assessment of whether a reasoning step is correct. In contrast, P2T is derived from a PRM explicitly trained to judge the correctness of reasoning steps, so its token-level credit is grounded in step-level correctness.

\begin{table}[t]
\centering
\caption{\label{tb:exp_compare_dense_reward}
Comparison of different dense rewards for RL. All the experiments use the Eurus-2-7B-SFT \citep{prime} as the baseline model, which is finetuned from Qwen2.5-Math-7B-Base on math and code datasets. The training dataset for RL is Eurus-2-RL-Data \citep{prime}.
}
\vspace{-0.1cm}
\setlength\tabcolsep{10pt}
\resizebox{0.98\textwidth}{!}{
\begin{tabular}{l|cccccc|c}
\toprule[1pt]
\textbf{Model} & \textbf{AMC23} & \textbf{MATH} & \textbf{OlympiadBench} & \textbf{K12} & \textbf{CodeForces} & \textbf{CodeContests}  & \textbf{Avg} \\
\midrule
Eurus-2-7B-SFT    
& 21.1 & 48.1 & 13.9 & 40.0 & 5.2  & 11.3 & 23.3 \\
\quad + GRPO    
& 23.6 & 51.8 & 20.5 & 48.6 & 28.6 & 28.1 & 33.5 \\
\quad + PRIME
& 31.2 & 52.7 & 25.4 & 54.6 & 26.4 & 25.9 & 36.0 \\
\quad + SPRO
& 31.9 & 53.6 & 28.2 & 55.0 & 29.4 & 32.1 & 38.4 \\
\rowcolor{exp_table_blue}
\quad + GRPO + P2T  
& \textbf{33.8} (\gainred{+1.9})  & \textbf{55.4} (\gainred{+1.8})  & \textbf{30.6} (\gainred{+2.4})  & \textbf{56.1} (\gainred{+1.1})  & \textbf{32.3} (\gainred{+2.9})  & \textbf{35.8} (\gainred{+3.7})  & \textbf{40.7} (\gainred{+2.3})  \\
\bottomrule[1pt]
\end{tabular}
}
\end{table}

\subsection{Ablation Study}
\label{sec:ablation}

\tabref{tb:exp_ablation} examines the key design choices of P2T. Unless otherwise noted, all experiments follow the Qwen2.5-Math-7B setting of \tabref{tb:exp_compare_outcome_llm} and report the average Pass@1 over the seven math benchmarks; the default configuration (53.8) thus matches the GRPO+P2T result in \tabref{tb:exp_compare_outcome_llm}. The exception is \tabref{tb:exp_ablation}(e), which follows the Qwen2.5-VL-7B-Instruct setting and reports the average over the five multimodal benchmarks of \figref{fig:exp_compare_outcome_mllm}. Each ablation varies a single factor while keeping all others at their defaults (shaded in gray).

\vspace{0.1cm}
\noindent\textbf{Token-level redistribution is the main source of the gains.}
\tabref{tb:exp_ablation}(a) decomposes the gain of P2T. Adding the step-level process reward to the outcome reward, \ie assigning each token the reward of its step (equivalent to $\eta=0$ in \eqref{eq:p2t_reward}), raises the accuracy from 48.0 to 50.6. Redistributing this reward among the tokens of each step with P2T further increases it to 53.8. Thus, of the overall gain of +5.8\%, +3.2\% stems from within-step credit assignment rather than from the step-level scores themselves. The benefit of P2T therefore cannot be explained merely by the use of an additional reward model: the same PRM signal becomes substantially more useful once it is attributed to the tokens responsible for it.

\vspace{0.1cm}
\noindent\textbf{P2T is robust to the redistribution strength $\eta$.}
As shown in \tabref{tb:exp_ablation}(b), the accuracy varies by at most 0.6\% as $\eta$ ranges from 0.25 to 1.0, and all settings remain well above the process reward baseline ($\eta=0$, 50.6). The exact strength of the redistribution therefore matters little once it is enabled, and $\eta$ requires no careful tuning. The slight decline for $\eta > 0.5$ is consistent with the role of $\mathcal{R}_s$ as a stabilizing baseline (\secref{sec:p2t_cal}): a larger $\eta$ places more weight on the attribution scores and hence amplifies their noise.

\vspace{0.1cm}
\noindent\textbf{Magnitude-based redistribution outperforms the signed variant.}
As shown in \tabref{tb:exp_ablation}(c), replacing the magnitude-based reward in \eqref{eq:p2t_reward} with its signed variant in \eqref{eq:p2t_reward_signed} lowers the accuracy from 53.8 to 52.6, although the signed variant still outperforms the process reward baseline (50.6). As noted in \secref{sec:p2t_cal}, a possible reason is that the positive and negative terms in \eqref{eq:p2t_reward_signed} are generally imbalanced within a step. Making better use of the sign information is left for future work.

\vspace{0.1cm}
\noindent\textbf{Better PRMs yield better token rewards.}
\tabref{tb:exp_ablation}(d) and (e) compare different PRMs. Among text-only PRMs, ReasonFlux-PRM-7B \citep{reasonflux-prm} performs best (53.8), followed by Qwen2.5-Math-PRM-7B \citep{prmlessons} (52.4) and Skywork-o1-Open-PRM-Qwen-2.5-7B \citep{skywork-o1-prm} (50.0). The same trend holds for multimodal PRMs, where VisualPRM \citep{visualprm} (57.5) outperforms MM-PRM \citep{du2025mmprmenhancingmultimodalmathematical} (55.3). With every PRM we tested, P2T improves over the outcome reward baseline (48.0 for the text-only model and 54.9 for the multimodal model), which shows that P2T works with a range of off-the-shelf PRMs and can directly benefit from stronger ones as they become available. At the same time, the gap between the best and worst text-only PRMs (3.8\%) is comparable to the gain of P2T itself, indicating that PRM quality has a substantial impact on P2T. In practice, if training becomes unstable, we therefore recommend first verifying that the PRM is competent on the training data, \eg that its step-level judgments are reliable for the target domain and response format.

\begin{table}[t]
    \centering
    \small 
    \caption{\label{tb:exp_ablation}
    Ablation Study for key designs in our P2T reward calculation. The experimental setup is identical to the Qwen2.5-Math-7B, except for (e), which follows the Qwen2.5-VL-7B-Instruct setup.}
    \vspace{-0.1cm}
    \begin{minipage}[t]{0.395\textwidth}
        \centering
        \footnotesize
        \subcaption*{(a)}
        \vspace{-0.1cm}
        \setlength\tabcolsep{5pt}
        \adjustbox{max width=\linewidth}{
        \begin{tabular}{c|c|c|c}
            \toprule
            \textbf{Reward}  & Outcome & \tabincell{c}{+ Process\\Reward}  & \tabincell{c}{+ P2T Token\\Reward}  \\
            \midrule
            Pass@1           & 48.0 & 50.6 & \cellcolor{gray}53.8  \\
            \bottomrule
        \end{tabular}
        }
    \end{minipage}\hfill
    \begin{minipage}[t]{0.295\textwidth}
        \centering
        \footnotesize
        \subcaption*{(b)}
        \vspace{-0.1cm}
        \setlength\tabcolsep{5pt}
        \adjustbox{max width=\linewidth}{
        \begin{tabular}{c|c|c|c|c}
            \toprule
            \textbf{\tabincell{c}{$\eta$ in\\\eqref{eq:p2t_reward}}}  & 0.25  & 0.5 & 0.75 & 1.0  \\
            \midrule
            Pass@1     & 53.5 & \cellcolor{gray}53.8  & 53.4 &  53.2  \\
            \bottomrule
        \end{tabular}
        }
    \end{minipage}\hfill
    \begin{minipage}[t]{0.282\textwidth}
        \centering
        \footnotesize
        \subcaption*{(c)}
        \vspace{-0.1cm}
        \setlength\tabcolsep{5pt}
        \adjustbox{max width=\linewidth}{
        \begin{tabular}{c|c|c}
            \toprule
            \textbf{\tabincell{c}{Reward\\Form}}  & \tabincell{c}{Signed\\in \eqref{eq:p2t_reward_signed}} & \tabincell{c}{Magnitude\\in \eqref{eq:p2t_reward}} \\
            \midrule
            Pass@1           & 52.6 & \cellcolor{gray}53.8  \\
            \bottomrule
        \end{tabular}
        }
    \end{minipage}
    \par\vspace{0.2cm}
    \begin{minipage}[t]{0.4\textwidth}
        \centering
        \footnotesize
        \subcaption*{(d)}
        \vspace{-0.1cm}
        \setlength\tabcolsep{5pt}
        \adjustbox{max width=\linewidth}{
        \begin{tabular}{c|c|c|c}
            \toprule
            \textbf{\tabincell{c}{PRM\\(text-only)}} & \tabincell{c}{ReasonFlux\\PRM-7B} & \tabincell{c}{Qwen2.5-Math\\PRM-7B} & \tabincell{c}{Skywork\\PRM-7B}  \\
            \midrule
            Pass@1  & \cellcolor{gray}53.8 & 52.4  &  50.0  \\
            \bottomrule
        \end{tabular}
        }
    \end{minipage}\hfill
    \begin{minipage}[t]{0.23\textwidth}
        \centering
        \footnotesize
        \subcaption*{(e)}
        \vspace{-0.1cm}
        \setlength\tabcolsep{5pt}
        \adjustbox{max width=\linewidth}{
        \begin{tabular}{c|c|c}
            \toprule
            \textbf{\tabincell{c}{PRM\\(multimodal)}}  & \tabincell{c}{Visual\\PRM}  & \tabincell{c}{MM\\PRM}  \\
            \midrule
            Pass@1           & \cellcolor{gray}57.5   & 55.3  \\
            \bottomrule
        \end{tabular}
        }
    \end{minipage}\hfill
    \begin{minipage}[t]{0.366\textwidth}
        \centering
        \footnotesize
        \subcaption*{(f)}
        \vspace{-0.1cm}
        \setlength\tabcolsep{5pt}
        \adjustbox{max width=\linewidth}{
        \begin{tabular}{c|c|c|c|c}
            \toprule
            \textbf{\tabincell{c}{Null\\Token}}  & \tabincell{c}{Pad\\Token}  & \tabincell{c}{EOS\\Token} & \tabincell{c}{Zero\\Embedding} & \tabincell{c}{Vocab\\Mean}  \\
            \midrule
            Pass@1     & \cellcolor{gray}53.8 & 50.3  & 51.4 &  52.5  \\
            \bottomrule
        \end{tabular}
        }
    \end{minipage}
\vspace{0.3cm}
\end{table}

\vspace{0.1cm}
\noindent\textbf{The null token should be neutral within the response.}
\tabref{tb:exp_ablation}(f) compares different choices of the null token $\mathbf{e}_\varnothing$. The PRM's padding token \texttt{<|endoftext|>} performs best (53.8), followed by the mean embedding of the PRM's vocabulary (52.5), the all-zero embedding (51.4), and the EOS token \texttt{<|im\_end|>} (50.3). We attribute these differences to how neutral each baseline is with respect to the PRM's judgment. Although \texttt{<|endoftext|>} is seen during pre-training, where it marks document boundaries, it almost never occurs within a reasoning response and thus carries little content-level meaning in this context. By contrast, \texttt{<|im\_end|>} marks the end of a response turn in the chat template and is closely tied to the completion of a response, which is precisely what the PRM is trained to judge. Substituting it for a token therefore injects an evaluation-related signal instead of removing information, resulting in the lowest accuracy. The all-zero embedding and the vocabulary mean carry no such signal, but both are artificial vectors that the PRM never encounters as inputs during training. Substituting them pushes the input off the training distribution, which may render the first-order estimate in \eqref{eq:taylor} less reliable.

\section{Related Work}
\label{sec:related_work} 

Reinforcement fine-tuning (RFT) \citep{ppo,deepseek-math} can substantially enhance the performance of large models \citep{o1, deepseek-r1, qwen3-vl, qwen2.5-vl, gemini-flash}, including their reasoning and generalization abilities \citep{chu2025sftmemorizesrlgeneralizes}. 
The RFT paradigm generally relies on two key components. The first is the reinforcement learning algorithm, which uses these rewards to compute advantages and drive policy updates, progressively refining the model’s parameters to maximize expected performance under the given reward criteria. The second is the reward function or model, responsible for assessing the quality and alignment of the model’s outputs and providing the scalar feedback that guides learning. Next, we introduce the specific designs of the reward functions.

\noindent\textbf{Outcome Reward.} 
Initially, a series of works starting with DeepSeek-R1  \citep{deepseek-r1} leveraged the simplest form of reward construction: a binary outcome reward that solely determines whether the model's final output is correct. This outcome reward paradigm has proven highly effective on numerous reasoning tasks, such as code and mathematical reasoning \citep{o1,qwen2.5,qwen2.5-coder,qwen3,gemini-flash,openr1-math-220k}. 
Recent work \citep{wang2024selftaughtevaluators,li2023generativejudgeevaluatingalignment,mcaleese2024llmcriticshelpcatch,zheng2023judgingllmasajudgemtbenchchatbot} involves using powerful Large Language Models (LLMs) as reward models. These LLMs are instructed to score the output of the model under training based on grammar, correctness relative to the truth, and preference. This score is then used as the reward. The advantage of this technique is its versatility, making it suitable for a wide range of unstructured tasks. 
Furthermore, several open-source reward models \citep{liu2025inferencetimescalinggeneralistreward,qwen2.5-math}, such as DeepSeek-GRM \citep{liu2025inferencetimescalinggeneralistreward}, are available and can be employed for reward modeling across a broad range of applications, including writing and mathematical proof. 
However, outcome rewards suffer from a severe limitation: reward sparsity. The vast majority of tokens and intermediate reasoning steps generated by the model do not receive customized reward signals. This often leads to training instability, where, for instance, incorrect early steps in an otherwise correct reasoning trajectory might still be inadvertently reinforced \citep{uesato2022solvingmathwordproblems}. 

\vspace{0.3cm}
\noindent\textbf{Process Reward.}
Due to the shortcomings of outcome rewards, research into process supervision \citep{uesato2022solvingmathwordproblems,li2023makinglargelanguagemodels,ma2023letsrewardstepstep,pan2023letsreinforcestepstep,wang2024mathshepherdverifyreinforcellms,luo2024improvemathematicalreasoninglanguage,lu2024autopsvautomatedprocesssupervisedverifier,li2025processrewardmodelqvalue,setlur2024rewardingprogressscalingautomated,liu2025adaptivestepautomaticallydividingreasoning,zeng2025versaprmmultidomainprocessreward}, which provides a more fine-grained reward signal than the outcome alone, has gained significant attention. 
Training a Process Reward Model (PRM) requires extensive annotation of fine-grained data. OpenAI \citep{lightman2023letsverifystepstep} utilized human labor to annotate a massive volume of step-level preference data specifically in the mathematics domain, which could then be used to train mathematics-focused PRMs. 
Math-Shepherd \citep{wang2024mathshepherdverifyreinforcellms} proposed an automated pipeline for efficiently annotating step-wise reward data. Specifically, it generates multiple rollouts starting from an arbitrary existing step, calculates the accuracy of these rollouts, and uses this accuracy value as an effective proxy for the reward signal of that particular step. 
Some open-source works \citep{deepseek-math}, such as qwen2.5-math-PRM \citep{prmlessons}, also provide options that can serve as PRMs for more general scenarios. 
Some recent works \citep{zhang2025gmprmgenerativemultimodalprocess,du2025mmprmenhancingmultimodalmathematical,visualprm,luo2025unlockingmultimodalmathematicalreasoning,liu2025divingselfevolvingtrainingmultimodal} have also explored constructing PRMs for multimodal settings. VisualPRM \citep{visualprm}, for instance, compiles 400K reward samples for vision-language PRM training and shows promising results in both multimodal and text-only reward tasks. Similarly, URSA \citep{luo2025unlockingmultimodalmathematicalreasoning}  collects over one million vision-language reward samples to train an 8B PRM, substantially boosting the capabilities of math-oriented VLLMs. 
Although process rewards are more fine-grained than outcome rewards, they remain relatively coarse, as each step still encompasses many tokens \citep{dapo}. 

\vspace{0.3cm}
\noindent\textbf{Token Reward.}  Constructing token-level rewards and appropriately assigning them to individual tokens can significantly improve both training effectiveness and efficiency \citep{prime,oreal,t-spmo,spro,scar,capo,tvm}. 
PRIME \citep{prime} proposes using the difference in log probabilities between the reward model and the reference model as the reward for each token. 
SPRO \citep{spro} assigns each token a reward based on the difference in log probabilities between the policy model and a reference model, without requiring any additional reward model. 
OREAL \citep{oreal} trains a token-level reward model by reusing the policy network backbone to assign scalar scores to each token, ensuring that the average token score matches the trajectory’s final binary reward, and uses these scores to reweight the policy loss during training. 
CAPO \citep{capo} converts step-level rewards into token-level rewards by assigning each token the reward of its corresponding step, using a pre-existing LLM to identify incorrect reasoning steps and applying a penalty to tokens within those steps, without requiring a dedicated token-level reward model. 
TVM \citep{tvm} trains a token-level reward model, built on the LLM architecture with an added scalar head, to predict each token’s probability of leading to a correct final answer, using sampled reasoning paths and minimizing the MSE between predicted scores and the true probabilities. 
T-SPMO \citep{t-spmo} assigns each token a reward equal to its expected incremental contribution given the preceding prefix, computed as the difference between the average rewards of responses with the prefix and those with the prefix followed by the token, relying solely on binary outcome rewards without any additional model.

\section{Conclusion}

In this paper, we present P2T, a training-free method that converts the coarse-grained reward of an existing process reward model into token-level rewards. P2T measures the influence of each token as the change in the process reward when the token is replaced with a null token, and efficiently estimates this influence for all tokens with a first-order Taylor approximation, which requires only a single forward and backward pass. The resulting token-level rewards can be readily integrated into standard RL algorithms such as GRPO, without training any additional reward model. Extensive experiments show that P2T consistently outperforms the outcome reward across model families, reasoning paradigms, and modalities, with particularly large gains where outcome-based RL stalls and on harder benchmarks. P2T further improves models that have already converged under outcome-based RL, reaches the best accuracy of outcome-based RL with 3--4$\times$ fewer training steps, and outperforms existing token-level reward methods.

\vspace{0.1cm}
\noindent\textbf{Limitations and future work.}
Since P2T derives its token-level rewards from a PRM, its effectiveness depends on the quality of the PRM, and a PRM that is unreliable on the target domain may limit its benefits. In future work, we will further explore how to exploit the sign of the token attribution scores, which indicates whether a token raises or lowers the process reward, to provide more informative token-level rewards.

\section{Acknowledgements}
This work was supported in part by the Research Grants Council under the Areas of Excellence scheme grant AoE/E-601/22-R.
The work has been supported by Hong Kong Research Grant Council-General Research Fund Scheme (Grant No. 17202422, 17212923, 17215025), Theme-based Research (Grant No. T45-701/22-R), and Strategic Topics Grant (Grant No. STG3/E-605/25-N). Part of the described research work is conducted in the JC STEM Lab of Robotics for Soft Materials funded by The Hong Kong Jockey Club Charities Trust.

{
    \small
    \bibliographystyle{unsrtnat}
    \bibliography{main}
}

\end{document}